\documentclass[runningheads]{llncs}
\usepackage[T1]{fontenc}
\usepackage{graphicx}
\usepackage{amsmath,amsfonts}
\usepackage{algorithmic}
\usepackage{algorithm}
\usepackage{array}
\usepackage{textcomp}
\usepackage{stfloats}
\usepackage{url}
\usepackage{verbatim}
\usepackage{graphicx}
\usepackage{cite}
\usepackage{multirow} 
\usepackage{float}
\graphicspath{ {image/} }

\usepackage{url}
\usepackage{makecell}
\usepackage{multirow}
\usepackage{rotating}
\usepackage{threeparttable} 
\usepackage{amsmath}  
\usepackage{amsfonts,amssymb}
\renewcommand{\algorithmicrequire}{\textitbf{Input:}}  
\renewcommand{\algorithmicensure}{\textitbf{Output:}} 

\usepackage{booktabs}
\usepackage{arydshln}
\usepackage[misc]{ifsym}
\usepackage{xcolor}

\begin{document}
\title{A Federated Parameter Aggregation Method for Node Classification Tasks with Different Graph Network Structures}
%
%
\author{
Hao Song\inst{1,2,3} \and
Jiacheng Yao\inst{1,2,3} \and
Zhengxi Li\inst{3} \and
Shaocong Xu\inst{1,2,3} \and
Shibo Jin\inst{1,2,3} \and
Jiajun Zhou\inst{1,2} \and
Chenbo Fu\inst{1,2} \and
Qi Xuan\inst{1,2} \and
Shanqing Yu\inst{1,2} \textsuperscript{(\Letter)}
}

\authorrunning{Song et al.}
%
\institute{
Institute of Cyberspace Security, Zhejiang University of Technology, \\Hangzhou 310023, China  
\and
Binjiang Cyberspace Security Institute of ZJUT, \\Hangzhou 310056, China
\and
College of Information Engineering, Zhejiang University of Technology, \\Hangzhou 310023, China \\
\email{yushanqing@zjut.edu.cn}}
%
\maketitle              
\begin{abstract}

Over the past few years, federated learning has become widely used in various classical machine learning fields because of its collaborative ability to train data from multiple sources without compromising privacy. However, in the area of graph neural networks, the nodes and network structures of graphs held by clients are different in many practical applications, and the aggregation method that directly shares model gradients cannot be directly applied to this scenario. Therefore, this work proposes a federated aggregation method FLGNN applied to various graph federation scenarios and investigates the aggregation effect of parameter sharing at each layer of the graph neural network model. The effectiveness of the federated aggregation method FLGNN is verified by experiments on real datasets. Additionally, for the privacy security of FLGNN, this paper designs membership inference attack experiments and differential privacy defense experiments. The results show that FLGNN performs good robustness, and the success rate of privacy theft is further reduced by adding differential privacy defense methods.

\end{abstract}

\section{Introduction}
The graph, consisting of nodes and edges, is a form of relational data that has attracted significant interest from researchers due to its wealth of structural information~\cite{posfai2016network,fu2011network,fu2019nes,xuan2021graph}. Among various models, graph neural network is widely recognized for performing complex tasks on graphs~\cite{gori2005new}, including node classification~\cite{gong2023neighborhood,xie2023pathmlp,zhou2024clarify}, graph classification~\cite{wang2021sampling,shen2021identity,zhou2022behavior,zhou2020m,zhou2020data}, and link prediction~\cite{zhang2018link,bhagat2011node,zhang2018end,8792200,8281007}. The remarkable performance has allowed it to be applied in various real-world scenarios~\cite{wu2020comprehensive}, such as semantic segmentation~\cite{guo2018view,yi2017syncspeccnn}, recommendation systems~\cite{berg2017graph,monti2017geometric,ying2018graph}, text categorization~\cite{ahmed2017inductive,kipf2016semi}, social effect prediction~\cite{qiu2018deepinf}, machine translation~\cite{luong2015effective,wu2016google}, and brain neural prediction~\cite{kawahara2017brainnetcnn}. However, a substantial amount of data is necessary for graph neural networks to achieve optimal performance. Inadequate training data hinder graph neural networks from achieving results comparable to those of other deep learning models.

Although multi-source datasets can solve a single dataset's low quality and improve accuracy, data security and privacy protection remain challenging. Federated learning is an effective method for processing isolated data by sharing gradient parameters to train a general global model while keeping the data local. It is often used in image processing applications such as target recognition and image classification~\cite{liu2020fedvision}. However, federated models for images cannot be directly applied to graph networks due to the two types of information in the latter: sample and structure. Commonly used federated learning methods for GNNs achieve federated aggregation by sharing node embedding vectors. For example, PPGNN proposed by chen et al.~\cite{chen2020vertically} used averaging, concatenation, and regression aggregation algorithms to aggregate shared node embeddings for node classification. Similarly, Ni~\cite{ni2021vertical} proposed a federated learning framework, FedVGCN, based on homomorphic encryption and secret sharing, which is used in graph convolutional networks. Both approaches focus on vertical federation environments, but in the case of horizontal federation scenarios, Zheng~\cite{zheng2021asfgnn} proposed an Automatic Separation Joint Graph Neural Network (ASFGNN) learning paradigm. This method optimizes hyperparameters using Bayesian methods to obtain better federated models and reduce communication burdens. Specific studies on applications of federated learning and graph neural networks in other fields have also been conducted. For instance, Wu et al.~\cite{wu2021fedgnn} developed the graph federated learning framework FedGNN based on recommender systems, while He et al.~\cite{lin2021fednlp} introduced a benchmark framework FDNLP for natural language processing to evaluate federated learning methods for different task formulations such as text classification, sequence tagging, question answering. However, these applications are limited to fixed fields and cannot be directly transferred to new scenes as data distribution between graph networks changes with the federated learning scene. In conclusion, there is still little research on federated aggregation methods for handling differences in client graph network structures in various horizontal federated learning scenarios.

To this end, this work proposes a parameter aggregation method, FLGNN, suitable for horizontal federated learning in various graph network scenarios to achieve node classification tasks. Based on the graph neural network, FLGNN uses the feature weight matrix as shared parameters, hides the feature vectors of nodes, and maintains excellent model performance while protecting private data. Moreover, in cases where the edge types of each client's network are not uniform, this paper proposes an aggregation strategy called FLGNN+, which can dynamically modify the aggregation weight based on model performance. 

The contributions of this work are as follows:
\begin{itemize}
\item: A GNN-based federated aggregation method for node classification is proposed in this study, which employs a sharing strategy based on multi-layer GNN weight parameters to adapt to scenarios with different client network structures. Experiments have demonstrated that in various horizontal federated learning scenarios, the global model obtained by this method is only about 1\%-2\% inferior to the model obtained by training the data together.

\item: This research discusses a dynamic aggregation strategy of feedback aggregation weights based on FLGNN, which is suitable for scenarios with different types of client network edges, and verifies its effectiveness on real datasets.

\item: A membership inference attack is designed in this paper to verify the privacy security of FLGNN. The experimental results show that FLGNN combined with the differential privacy defense method can reduce the success rate of being stolen to 30\%-50\% of the training alone.
\end{itemize}

The rest of this paper is organized as follows: Section~\ref{related work} reviews related work on graph neural networks. Section~\ref{the proposed model} describes the process and principle of FLGNN and FLGNN+ methods in detail. Section~\ref{experiments} presents the relevant experimental settings. Section~\ref{conclusion} provides a summary of the methodology and directions for future research.

\section{ Related Work}
\label{related work}
\subsection{Graph Neural Network}
Graph Attention Network (GAT) ~\cite{velickovic2017graph} is employed as the graph neural network model in this study. The core of GAT lies in the attention mechanism, which applies distinct weights to each of the target node's neighbors based on the relevance of the nodes. By doing so, GAT can avoid relying on specific network topologies and achieve superior results. This unique capability makes it suitable for a wide range of applications, including recommendation systems~\cite{song2019session} and trajectory prediction~\cite{kosaraju2019social}. The attention cross-correlation index $e_{ij} = a(Wh_i, Wh_j)$ indicates the significance of the first-order neighbor $j$ of node $i$. In Eq.~\ref{eq1}, $W$ stands for a trainable feature weight matrix, and $h$ stands for the input node's feature vector. $\mathcal{N}_i$ is the collection of first-order neighbors of node $i$, and $\alpha_{ij}$ is the introduction of softmax to regularize the attention cross-correlation index of all first-order neighbors of node $i$.
\begin{equation}\label{eq1}
\centering
\alpha_{ij} = \frac{exp(Leakyrelu(\vec{a}^{T}[Wh_i||Wh_j]))}{\sum_{k\in \mathcal{N}_i}exp(Leakyrelu(\vec{a}^{T}[Wh_i||Wh_k]))}
\end{equation}

\subsection{Privacy Attack and Defense}
During federated learning training, clients' privacy may be at risk of being leaked or stolen. Membership inference attack~\cite{jia2019memguard}, graph reconstruction attack~\cite{aono2017privacy}, and attribute inference attack~\cite{narayanan2008robust} are common techniques used to compromise privacy in graph networks. Depending on whether the adversary can access intermediate computation results or model parameters, these attacks can be classified into black-box and white-box attacks. This paper uses the confidence attack to infer client membership, where the attacker leverages the principle that a node with high confidence is likely to be the training node of the target client~\cite{duddu2020quantifying}. The adversary typically determines the optimal confidence threshold to distinguish training samples from non-training samples by analyzing statistical variations in the output predictions of distinct samples.
\hspace*{1em}Differential privacy noise is typical mitigation against member inference attacks. Adding Gaussian, Laplace, and binomial noise are common approaches for achieving differential privacy. This paper uses laplace noise, and the noise $Y\sim L(0,\Delta f/\epsilon)$ must meet the $(\epsilon,0)$ condition. For a given arbitrary domain function $M: \mathcal{D}\rightarrow \mathcal{R}^d$, any input $x$ can be added noise according to Eq.~\ref{eq3}~\cite{dwork2006calibrating,choudhury2019differential}, where $\Delta f$ stands for sensitivity, $\left\|*\right\|$ for the vector's norm, $Lap(b)$ for the parameter $b$ conforms to the Laplacian distribution, $f(*)$ for the query function, and $D$ and $D'$ for two datasets with only one data difference, $\epsilon$ reflects the budget for privacy. In general, the lower the privacy budget, the higher the noise and the stronger the privacy protection impact, but the model's performance will suffer.
\begin{equation}\label{eq2}
\centering
\Delta f = \max_{D,D'}{\left\|f(D)-f(D')\right\|}
\end{equation}

\begin{equation}\label{eq3}
\centering
M(x)+Lap(\Delta f/\epsilon)
\end{equation}
\hspace*{1em}This paper considers a semi-honest scenario ~\cite{mohassel2017secureml,hardy2017private} where both the client and the server strictly adhere to the security protocol. Each data holder is only aware of their own data and is unaware of other data holders' local data. Furthermore, the server will not collude with any data holders. Curious clients may try to infer as much information as possible about other clients from intermediate results.

\section{The Proposed Model}
\label{the proposed model}
\subsection{Problem Formulation}
\begin{figure}[h] 
    \centering 
    \includegraphics[width=1\textwidth]{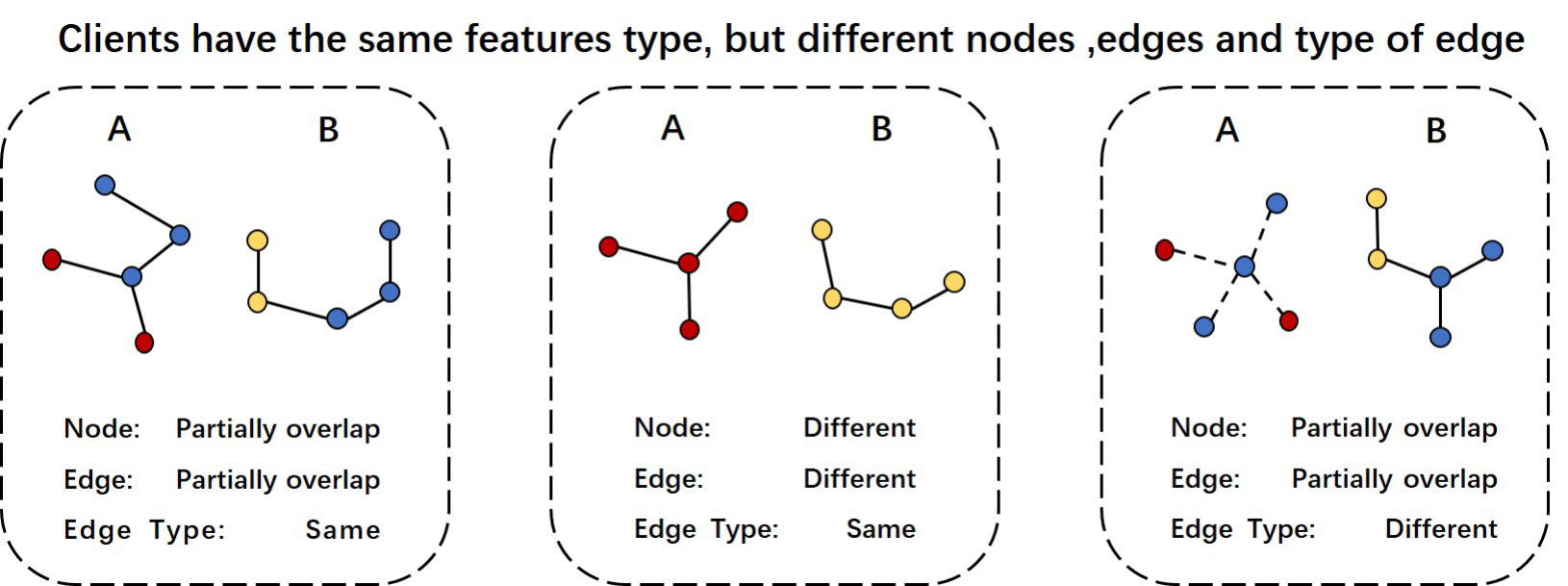} 
    \caption{Three federation situations are depicted in the diagram. Two clients' networks are represented by A and B. The blue nodes show common nodes between the two networks, while the red and yellow nodes represent network-specific nodes. Different sorts of edges are shown by the dotted and solid lines.} 
\label{fig1}
\end{figure}

This paper investigates the challenges of node classification in a semi-supervised horizontal federation scenario. Figure~\ref{fig1} illustrates three client-held graph network scenarios, including partially overlapping nodes and edges, completely different nodes and edges, and different types of edges. Assume there are $N$ clients, each of whom has a portion of the data on the whole graph. Figure~\ref{fig1} depicts a case where there are two clients, each of whom possesses a portion of the data on the entire graph. $V$ and $E$ are defined as the sets of nodes and edges in the entire graph network, and $V_u$, $E_u$ denote the sets of nodes and edges in any given client, where $\bigcup_{1}^{N}V_u=V$, $\bigcup_{1}^{N}E_u=E$, $u\in{1, 2, …, N}$. In the extreme scenario where there is no overlap between nodes and edges, $\bigcap_{1}^{N}V_u=\emptyset$, $\bigcap_{1}^{N}E_u=\emptyset$. Moreover, in cases where there is a disparity in feature dimensions between nodes, alignment of the feature dimensions can be accomplished by augmenting the missing feature column and padding it with zeros. $G_u(A_u, X_u)$ is the graph kept by the client $u$, where $A_u$ is the node's adjacency matrix, $a_{ij}^u$ is an element in the adjacency matrix, $a_{ij}^u=1$ denotes node i and node j are linked, otherwise $a_{ij}^u=0$; $X_u$ is the node's feature matrix, $X_u\in R^{k\times n}$, where $k$ is the number of nodes in client $u$ and $n$ is the dimension of the node feature.

\subsection{Model Training And Aggregating}

\begin{figure*}[h]
    \centering 
    \includegraphics[width=1\textwidth]{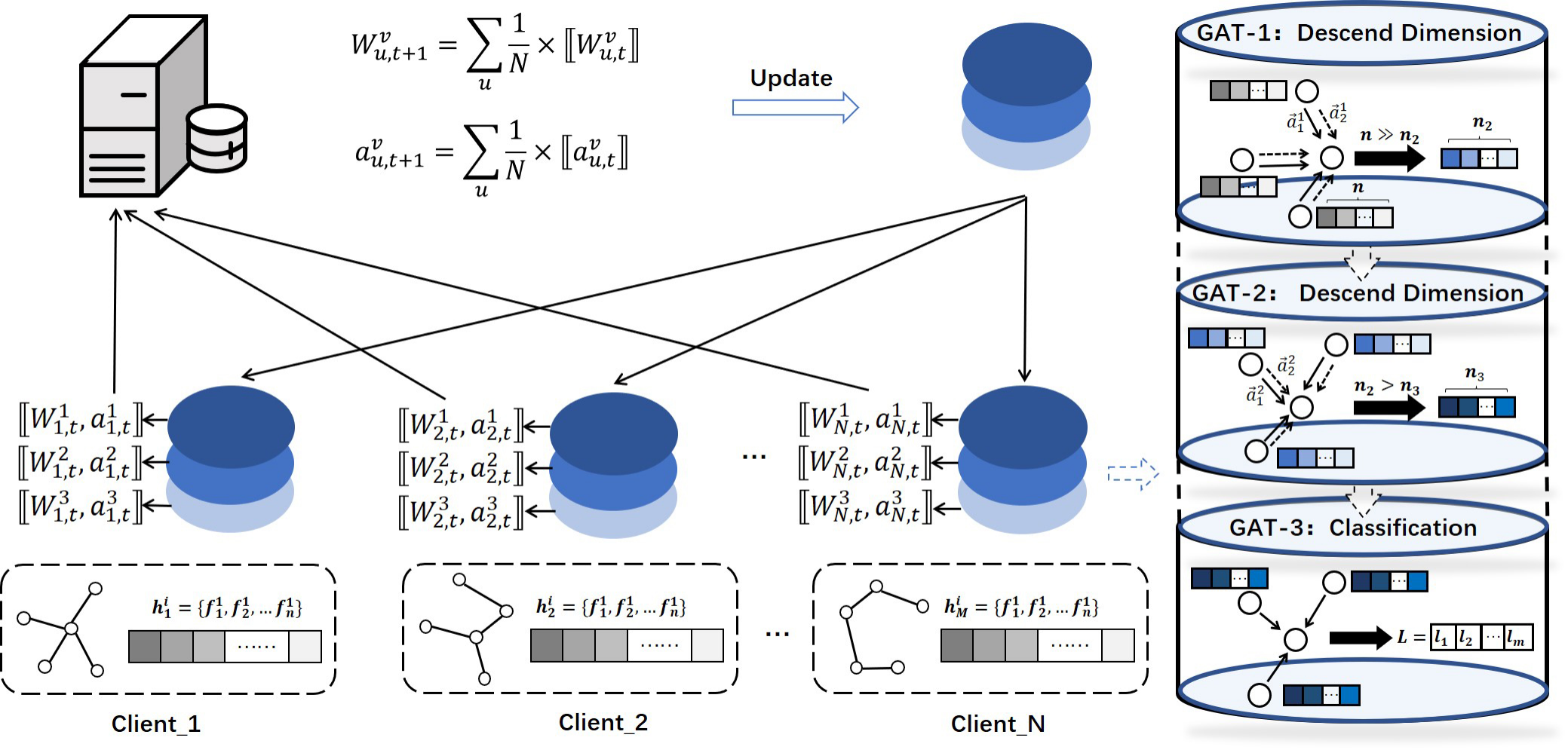}
    \caption{The left side of the figure depicts the process of federated training and aggregation, where $[\![*]\!]$ represents the encrypted parameter, $i$ is the number of graph neural network layers, and $n,n_2,n_3$ is the number of feature dimension. The right side of the figure depicts FLGNN specific structure. Furthermore, $m$ represents the number of categories, and $\vec{a}$ represents the head of the graph attention mechanism.}
\label{fig2} 
\end{figure*}

\begin{figure}[h]
    \centering
    \includegraphics[width=1\textwidth]{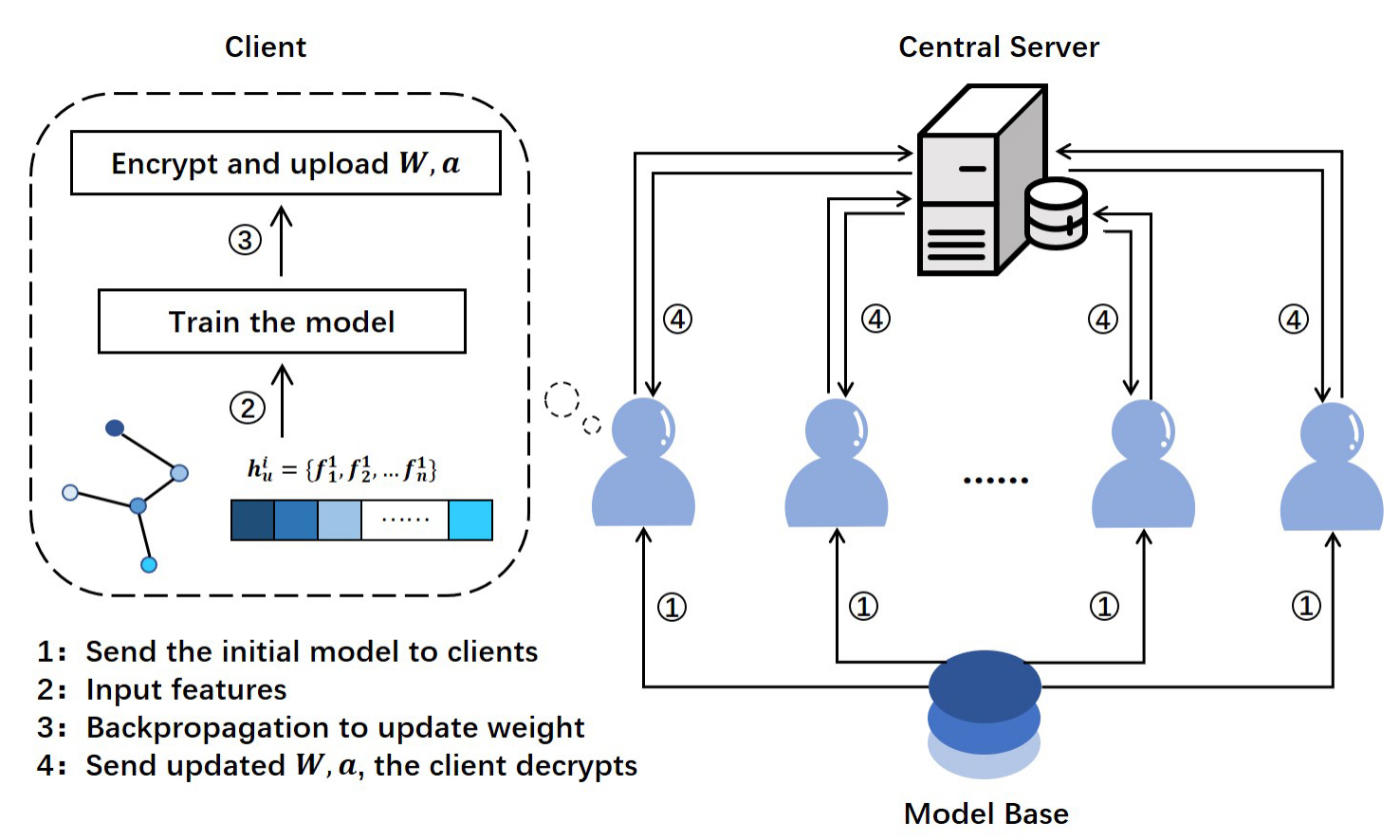}
    \centering
    \caption{Federal training flow chart.}
\label{fig3}
\end{figure}

Figure~\ref{fig2} details the FLGNN procedure, which requires trustworthy third-party servers and a large number of data holders as clients to participate in federated learning. After a certain batch of local training, the client encrypts and uploads the obtained model weight parameters to the server. The aggregation strategy employed is federated averaging aggregation ~\cite{konevcny2016federated}. The server calculates the average value of the encrypted parameters and sends it to each client. The client updates the local model after decryption and begins the next epoch of training until the global model converges. The full training process is depicted in Algorithm 1, which is divided into two steps: local training and federated aggregation.

Figure~\ref{fig3} depicts the weight parameter aggregation process and the role of the three-layer GNN. The first and second layers employ a multi-head attention mechanism to aggregate the features of neighboring nodes into their own embedding vectors. The third layer serves as the output layer, which translates features into the same dimensions as labels for categorization. A multi-headed attention mechanism is not required for this layer. The model's aggregation is performed on a trustworthy central server, which may selectively aggregate the weight parameters of a single layer or several layers, and the impact of combining the weight parameters of different layers varies. Since different features have varying degrees of association with labels, some features closely related to labels are likely to be assigned higher weights during iterative training. It is widely believed that the first layer model contains the most comprehensive node features, so aggregating the weight parameters of this layer can learn the most information about nodes. However, the information contained in the original features after compression and fusion becomes redundant due to the reduced feature dimensions in the second and third layers. Therefore, aggregating the weight parameters of these two layers may yield inferior results compared to the first layer. Generally, aggregating more weight parameters can learn more information about other clients. Moreover, regardless of whether single-layer or multi-layer weight parameters are aggregated, the server can only aggregate the weight parameters of the same layer, but not across different layers.

\begin{algorithm}[t]
  \renewcommand{\algorithmicrequire}{\textbf{Input:}}
  \renewcommand{\algorithmicensure}{\textbf{Output:}}
  \caption{FLGNN}
  \label{alg:1}
  \begin{algorithmic}[1]
    \REQUIRE $h_u^i=\{f_1^1,f_2^1,...,f_n^1\}$,$i\in V$, Feature of the input node\\
             $W_{u}^v$,$\vec{a}_{u}^v$, $v\in\{1,2,3\}$, weight matrices and attention, random initialization;\\$\mathcal{N}_u^i$, The set of neighbor nodes of the target node $i$;\\$q$, aggregation frequency;\\$t$, iterative batch.
    \ENSURE Global model $M$
	\FOR {$t <= max\_epoch$}
	\STATE \#CG1:Local training
    \FOR {$v =1\to 3$}  
    \STATE Calculates attention coefficient\\$e_{ij}=a(W_{u}^vh_u^i,W_{u}^vh_u^j)$;$a()$ is a custom function
    \STATE Regularize adjacent nodes by formula (1) $\Rightarrow \alpha_{ij}$
    \STATE Calculate the output features of layer $v$\\$h_u^{i'}=\sigma(\sum_{j\in M_u^i}\alpha_{ij}*W_{u}^vh_u^j)$
	\IF {$v < 3$}
		\STATE Concat for each node $h_u^{i'}\Rightarrow H_u^{v'}$
    \STATE The output feature is used as the input of the next layer $H_u^{v'}\Rightarrow H_u^{v+1}$
	\ENDIF
    \STATE  $v=v+1$
    \ENDFOR
    \STATE Backpropagation to update weight parameter $W_{u,t}^v$, $a_{u,t}^v$
    \STATE \textbf{return} weight parameter $W_{u,t}^v$, $a_{u,t}^v$ for each client
	\STATE \#CG2:The server performs federated aggregation
	\IF {$t \% q == 0$}
	\STATE Perform aggregation $W_{u,t+1}^{v} = \sum_{u}\frac{1}{N} * W_{u,t}^v$,\\ $a_{u,t+1}^{v} = \sum_{u}\frac{1}{N} * a_{u,t}^v$
	\ENDIF
	\STATE $t = t + 1$
	\STATE \textbf{return} Global model for each client
	\ENDFOR
  \end{algorithmic}  
\end{algorithm}
\subsection{FLGNN+}
FLGNN+ proposes a new aggregation method for scenarios in which each client has completely different types of edges. The training process of FLGNN+ is the same as that of FLGNN. When clients participating in federated learning have different types of graph network connections, simply using the federated averaging aggregation algorithm will result in poor model performance due to the generalization of networks with varied connection types. Different types of edges have different impacts on the final task. For instance, in the task of detecting phishing accounts, transaction networks (edges indicate transaction relationships) and kinship networks (edges indicate kinship between accounts) of accounts play different roles in identifying malicious accounts. To deal with this problem, FLGNN+ assigns distinct aggregate weights $\gamma_u^n$ to each client and dynamically modifies them based on model accuracy feedback. FLGNN+ will be more prone to learn its weight parameters during dynamic adjustment since the local model cannot estimate the effect of other types of networks on the final task. The aggregation weight needs to satisfy $\sum_{N}\gamma_u^N =1$ to ensure the convergence of the model. The weight parameter update formula and weight adjustment formula are as follows, $W_{u,t}$, $a_{u,t}$ represent the weight parameter value and attention value of the client $u$ in the $t$-th training batch, which its initial value is $\frac{1}{N}$, and $\mu_{u,t}$ is adjustment factor. $M_{u,t}$ is the model accuracy of client $u$ in the $t$-th training batch. $L_{up}$ is the upper bound of the aggregation weight, and $\gamma_{u,t}^n$ is the aggregate weight in the $t$-th training batch of client. $\eta$ is a hyperparameter, indicating the adjustment range of each batch of model parameters
\begin{equation} \label{eq4}
\centering
W_{u,t+1} = \sum_{n}\gamma_{u,t}^n * W_{u,t}
\end{equation}

\begin{equation} \label{eq5}
\centering
a_{u,t+1} = \sum_{n}\gamma_{u,t}^n * a_{u,t}
\end{equation}

\begin{equation} \label{eq6}
\mu_{u,t}=\left\{
\begin{array}{rcl}
1 & & {M_{u,t+1} - M_{u,t} \le 0 and \gamma_{u,t}^n < L_{up}}\\
0 & & {else}\\
\end{array} \right.
\end{equation}

\begin{equation} \label{eq7}
\gamma_{u,t+1}^n=\left\{
\begin{array}{rcl}
\gamma_{u,t}^n+\mu_{u,t}*\eta & & {n=u}\\
1-\gamma_{u,t+1}^u /(n-1) & & {n \ne u}\\
\end{array} \right.
\end{equation}

\section{Experiments}
\label{experiments}
\subsection{Datesets And Experiments Setting}
We consider a set of graph datasets: Cora, Citeseer, Wiki, LastFM Asia, Terrorist Attack, and Yelp Urbana. The first three are citation network datasets ~\cite{kipf2016semi,yang2015network}, where each node represents a paper or document and each edge represents a citation relationship between documents. LastFM Asia is a social network dataset ~\cite{rozemberczki2020characteristic}, where nodes represent users of the music streaming media LastFM, and links represent mutual attention relationships between users. Terrorist Attack ~\cite{zhao2006entity}  is a network dataset on terrorist attacks published on the PIT (Profile In Terror) website, tracking six different types of terrorist assaults. The PIT website has compiled two networks: loc and loc org. The former refers to terrorist attacks that occurred at the same location. The latter pertains to a terrorist attack executed by the same organization at the same location. Yelp Urbana is a sub-dataset extracted from the public dataset maintained by the Yelp review site ~\cite{luca2016reviews}. Specifically, this research extracts Urbana's restaurant reviews from the Yelp review site to construct two networks: the friend network Yelp Urbana friends and the co-dining network Yelp Urbana co-dining. Table~\ref{table1} presents the specific counts of nodes, edges, features, and labels for these datasets.

The first four datasets are separated and given to the client in this article to simulate a federation scenario, while the latter two datasets are used for scenarios with networks with different edge types. Unless otherwise stated, the average accuracy achieved by 10-fold cross-validation is used as the indicator of the assessment model in this article. As for hyperparameters in this chapter's experiment, the learning rate $lr$ is set at 0.005, the $L_2$ regularization coefficient is 0.0005, the number of hidden layers $nhid$ is 8, the number of attention heads $nhead$ is 8, and the aggregation frequency $p = 2$.

In addition, the experimental environment in this chapter uses a single device to simulate federated multi-party communication. The specific device configuration is as follows: the host CPU model is Intel(R) Core(TM) i7-10710U, the main frequency is 1.1-4.7GHz, the number of cores is 6, and the memory is 32G. The GPU model is NVIDIA TESLA V100, and the video memory is 16G.
\begin{table}[]
\centering
\setlength{\tabcolsep}{2mm}
  \renewcommand\arraystretch{1.2}
\begin{tabular}{ccccc}
\hline
Dataset       & Node   & Edge &  Feature  & class \\ 
\hline
Cora  & 2708 & 5429 & 1433 & 7  \\
Citeseer & 3327 & 4732 & 3703 & 6  \\
Wiki & 2405 & 17981 & 4973 & 19  \\
LastFM Asia & 7624 & 27806 & 7842 & 18  \\
Yelp Urbana\_friends & 1872 & 3036 & 149 & 5 \\
Yelp Urbana\_co-dining & 2622 & 361901 & 149 & 5 \\
Terrorist Attack\_loc & 645 & 3172 & 106 & 6 \\
Terrorist Attack\_loc\_org & 260 & 571 & 106 & 6 \\
\hline
\end{tabular}
\label{table1}
\end{table}

\subsection{Aggregation Validation Experiment}
In order to verify the impact of aggregating weight parameters of different layers on the performance of the federated learning global model, this section designs FLGNN multi-layer aggregation experiments. Validation experiments on the first four datasets are carried out in this part. The dataset is randomly partitioned among different clients under the premise of ensuring relatively uniform labels. Following segmentation, each client must be ensured that nodes and edges only partially overlap and that the graph networks' magnitude stays roughly the same. Cora and Citeseer are separated into two clients, while Wiki and LastFM Asia are separated into five clients due to the large number of nodes in these two datasets. The dataset is divided into a training set, validation set, and test set in the ratio 1:2:7, and follow-up experiments comply with this setting if there is no special instruction. Table~\ref{table2} shows the experimental results. FLGNN\_L* represents the result obtained by the clients' model aggregating the parameters of the certain layer or multi-layer and testing on the corresponding test set; Alone represents the result obtained by the same model (trained separately by each client using their own train set) testing on each test set; Full represents the result obtained by the model (trained by using the entire dataset) testing on each client’s test set; A, B,..., E represents various clients.

The results show that there is a significant improvement over training alone by adopting FLGNN. Although the model's accuracy has dropped if compared to the training using the entire dataset, the decrease is much smaller than that of the single client's accuracy. Furthermore, the weight parameters of the first layer may increase the accuracy of the global model the most in most cases, and the effect of aggregating the weight parameters of the second and third layers is smaller than that of the first layer for a single layer, and sometimes even better than the effect of two-layer aggregating. This finding backs up the previous analysis. This is because the first layer's parameters are computed directly from the original features, and the parameters' dimensions are higher than the total of the other two layers' dimensions. The overall best aggregation strategy for multi-layer aggregation is to combine the weight parameters of the three layers, and the experiments that follow will employ a three-layer weight parameter aggregation strategy. Furthermore, the communication cost required to aggregate various layers differs, and the feature dimensions of the first layer will almost certainly increase the quantity of data transmitted. Aggregation of the second and third layers is also a smart choice when the communication bandwidth of the client participating in federated learning is restricted and the performance requirements of the model are not severe.\\

\begin{table*}[]
\centering
\setlength{\tabcolsep}{1.2mm}
\caption{Experimental Results}
\resizebox{1.0\linewidth}{!}{
\begin{tabular}{cccccllllllllll}
\hline
\multicolumn{1}{l}{} & \multicolumn{2}{c}{Cora}                                      & \multicolumn{2}{c}{Citeseer}                                  & \multicolumn{5}{c}{Wiki}                                                                                                                                              & \multicolumn{5}{c}{LastFM Asia}                                                                                      \\ \cline{2-15} 
                     & A                             & B                             & A                             & B                             & \multicolumn{1}{c}{A}                             & \multicolumn{1}{c}{B}      & \multicolumn{1}{c}{C}      & \multicolumn{1}{c}{D}      & \multicolumn{1}{c}{E}      & \multicolumn{1}{c}{A} & \multicolumn{1}{c}{B} & \multicolumn{1}{c}{C} & \multicolumn{1}{c}{D} & \multicolumn{1}{c}{E} \\ \hline
FLGNN\_L1               & 0.7909                        & 0.7633                        & 0.6243                        & 0.6253                        & \multicolumn{1}{c}{0.6314} & \multicolumn{1}{c}{0.6148} & \multicolumn{1}{c}{0.6274} & \multicolumn{1}{c}{0.6580} & \multicolumn{1}{c}{0.6260} & 0.8046                       & 0.8107                      & 0.8314                      & 0.8253                      & 0.8197                      \\
FLGNN\_L2               & 0.7849                        & 0.7381                        & 0.5895                        & 0.5871                        & \multicolumn{1}{c}{0.6244}                        & \multicolumn{1}{c}{0.5856} & \multicolumn{1}{c}{0.6145} & \multicolumn{1}{c}{0.6263} & \multicolumn{1}{c}{0.6041} & 0.8048                      & 0.8100                      & 0.8251                      & 0.8217                      & 0.8217                      \\
FLGNN\_L3               & 0.7832                        & 0.7387                        & 0.5980                        & 0.5909                        & \multicolumn{1}{c}{0.6271}                        & \multicolumn{1}{c}{0.5999} & \multicolumn{1}{c}{0.6195} & \multicolumn{1}{c}{0.6331} & \multicolumn{1}{c}{0.6117} & 0.7892                      & 0.8062                      & 0.8191                      & 0.8171                      & 0.8142                      \\
FLGNN\_L12              & \textbf{0.8009}                        & \textbf{0.7647}                        & \textbf{0.6302}                        & 0.6305                        & \multicolumn{1}{c}{0.6307}                        & \multicolumn{1}{c}{0.6083} & \multicolumn{1}{c}{0.6228} & \multicolumn{1}{c}{0.6572} & \multicolumn{1}{c}{0.6278}                            & 0.7979                      & 0.8126                      & 0.8306                      & 0.8228                      & 0.8193                      \\
FLGNN\_L13              & 0.7965                        & 0.7622                        & 0.6300                        & 0.6289                        & \multicolumn{1}{c}{0.6436}                        & \multicolumn{1}{c}{0.6268} & \multicolumn{1}{c}{0.6415} & \multicolumn{1}{c}{0.6648} & \multicolumn{1}{c}{0.6400}                            & 0.8078                      & 0.8167                      & 0.8328                      & 0.8285                      & 0.8239                      \\
FLGNN\_L23              & 0.7880                        & 0.7364                        & 0.5945                        & 0.5875                        & \multicolumn{1}{c}{0.6294}                        & \multicolumn{1}{c}{0.6010} & \multicolumn{1}{c}{0.6228} & \multicolumn{1}{c}{0.6431} & \multicolumn{1}{c}{0.6106}                            & 0.8002                      & 0.8089                      & 0.8236                      & 0.8160                      & 0.8111                      \\
FLGNN\_L123             & 0.7990                        & 0.7640                        & 0.6295                        & \textbf{0.6321}                        & \multicolumn{1}{c}{\textbf{0.6475}}                        & \multicolumn{1}{c}{\textbf{0.6266}} & \multicolumn{1}{c}{\textbf{0.6446}} & \multicolumn{1}{c}{\textbf{0.6714}} & \multicolumn{1}{c}{\textbf{0.6489}}                            & \textbf{0.8086}                      & \textbf{0.8177}                      & \textbf{0.8341}                      & \textbf{0.8291}                      & \textbf{0.8264}                      \\ \hline
Alone                & 0.7836                        & 0.7380                        & 0.5898                        & 0.5895                        & \multicolumn{1}{c}{0.6106}                        & \multicolumn{1}{c}{0.5792} & \multicolumn{1}{c}{0.6150} & \multicolumn{1}{c}{0.6264} & \multicolumn{1}{c}{0.6008}                            & 0.7992                      & 0.8024                      & 0.8141                      & 0.8127                      & 0.8091                      \\
Full            & 0.8131                        & 0.7651                        & 0.6390                        & 0.6525                        & \multicolumn{1}{c}{0.6572}                        & \multicolumn{1}{c}{0.6411} & \multicolumn{1}{c}{0.6504} & \multicolumn{1}{c}{0.6826} & \multicolumn{1}{c}{0.6573}                            & 0.8166                      & 0.8272                      & 0.8416                      & 0.8331                      & 0.8350                      \\\hline
\end{tabular}}
\label{table2}
\end{table*}

\subsection{Experiment of Nodes and Edges Without Repetition}
In this experiment, we focus on an extreme example in which the client's nodes and edges are entirely different. Moreover, in this situation, we guarantee that the total number of nodes on each client equals that of the original dataset when dividing the four datasets. Without repetition, segmentation will undoubtedly lose certain edges, resulting in a total number of edges for each client that is less than that of the original dataset, lowering the accuracy of the Full. The training set, validation set, and test set of the divided Wiki dataset are divided into 2:2:6 in this experiment because the training set of the divided Wiki dataset is tiny and the training set cannot include all of the labels sometimes. Table~\ref{table3} illustrates the experimental results, where FLGNN indicates an aggregation approach that employs the aggregation of three-layer weight parameters.

\begin{figure}[]
    \centering 
    \includegraphics[width=1.0\textwidth]{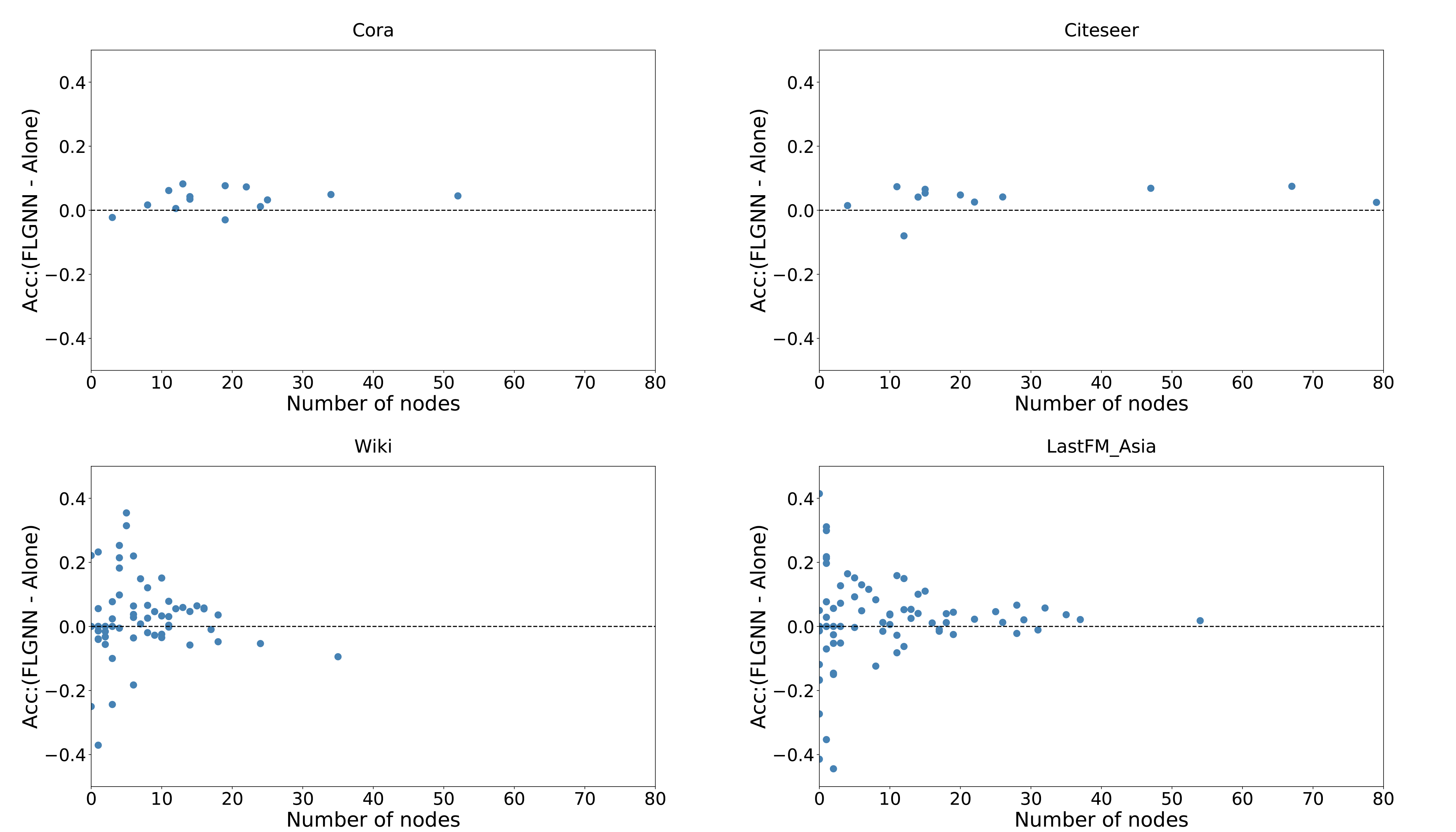} 
    \caption{Trend chart of the influence of the number of clients on federated aggregation. The abscissa is the number of clients, and the ordinate is the average accuracy of Full\_Client minus individual training and minus FLGNN}
\label{fig4}
\end{figure}

The results present that FLGNN continues to play a role in this severe condition, and the performance of its trained model outperforms that of independent training, particularly on the two datasets of Cora and Citeseer. On a client of Cora, the impact of FLGNN even outperforms Full. On Wiki and LastFM Asia, however, some clients may not have a significant improvement through federated learning. We then use a scatter plot to visualize the label prediction accuracy in each dataset to find the reasons for the unsatisfactory federal effect. Figure~\ref{fig4}. shows that the label distribution of the Cora and Citeseer datasets is fairly uniform, and FLGNN essentially increases the prediction accuracy of various labels. However, the distribution of labels is not uniform for the two datasets, Wiki and LastFM Asia, and some labels contain only one or two nodes or even no node at all. The paucity of nodes makes learning the right weight distribution of the label difficult, and the prediction accuracy is very sensitive to the prediction accuracy of a single node, which results in a great fluctuation. Therefore, only when there is a sufficient number of nodes participating in training, the effect of FLGNN gradually stabilizes. And the effect of FLGNN varies substantially when the label distribution of the data is exceedingly unequal.

\begin{table}[]
\centering
\setlength{\tabcolsep}{3.2mm}
\caption{Experimental Results}
\begin{tabular}{cccccc}
\hline
                              & Client & FLGNN  & Alone  & Full \\ \hline
\multirow{2}{*}{Cora}         & A      & 0.7966 & 0.7589 & \textbf{0.8054}   \\
                              & B      & \textbf{0.7676} & 0.7137 & 0.7509    \\ \hline
\multirow{2}{*}{Citeseer}     & A      & 0.6010 & 0.5519 & \textbf{0.6115}    \\
                              & B      & 0.6345 & 0.5874 & \textbf{0.6441}    \\ \hline
\multirow{5}{*}{Wiki}         & A      & 0.5699 & 0.5675 & \textbf{0.6119}    \\
                              & B      & 0.6418 & 0.6179 & \textbf{0.6655}    \\
                              & C      & 0.5769 & 0.5441 & \textbf{0.6150}    \\
                              & D      & 0.6551 & 0.6151 & \textbf{0.6775}    \\
                              & E      & 0.6642 & 0.6359 & \textbf{0.6846}    \\ \hline
\multirow{5}{*}{LastFM Asia}       & A      & 0.7646 & 0.7508 & \textbf{0.7782}    \\
                              & B      & 0.7721 & 0.7374 & \textbf{0.7859}    \\
                              & C      & 0.7577 & 0.7139 & \textbf{0.7718}    \\
                              & D      & 0.7649 & 0.7423 & \textbf{0.7821}    \\
                              & E      & 0.7366 & 0.7079 & \textbf{0.7484}    \\ \hline
\end{tabular}
\label{table3}
\end{table}

\subsection{Client Quantity Sensitivity Experiment}
The number of federated entities participating in real-world scenarios varies based on the complexity of the task. Therefore, this paper conducts a sensitive experiment based on the number of clients to evaluate the effectiveness of FLGNN across diverse application scenarios. Considering the network scale of the Wiki and LastFM Asia datasets, this article divides them into 2, 5, 7, and 10 clients. This division is based on the understanding that when the number of clients is small, a single client may train a superior model utilizing rich edge information.

The number of nodes and connections held by a single client reduces as the number of clients rises, and the accuracy of a single client training also drops dramatically, but FLGNN can still play a role in coordinating the classification of each client, as shown in Table~\ref{table4}. This result shows that the number of clients does not affect the effectiveness of FLGNN. In actuality, a region's data is always limited. Through FLGNN, dozens of firms with little data may attain the accuracy of federated learning with several industry giants. Furthermore, the data quality of the data possessor differs. The performance of each client's separately trained models varies due to the randomization of the data distribution, yet participation in the federation will be required. Clients with high-quality data require information from edge nodes in other client networks to complement their data, while those with low-quality data need more node information to increase model performance dramatically.

\begin{figure}[t] 
    \centering 
    \includegraphics[width=1.0\textwidth]{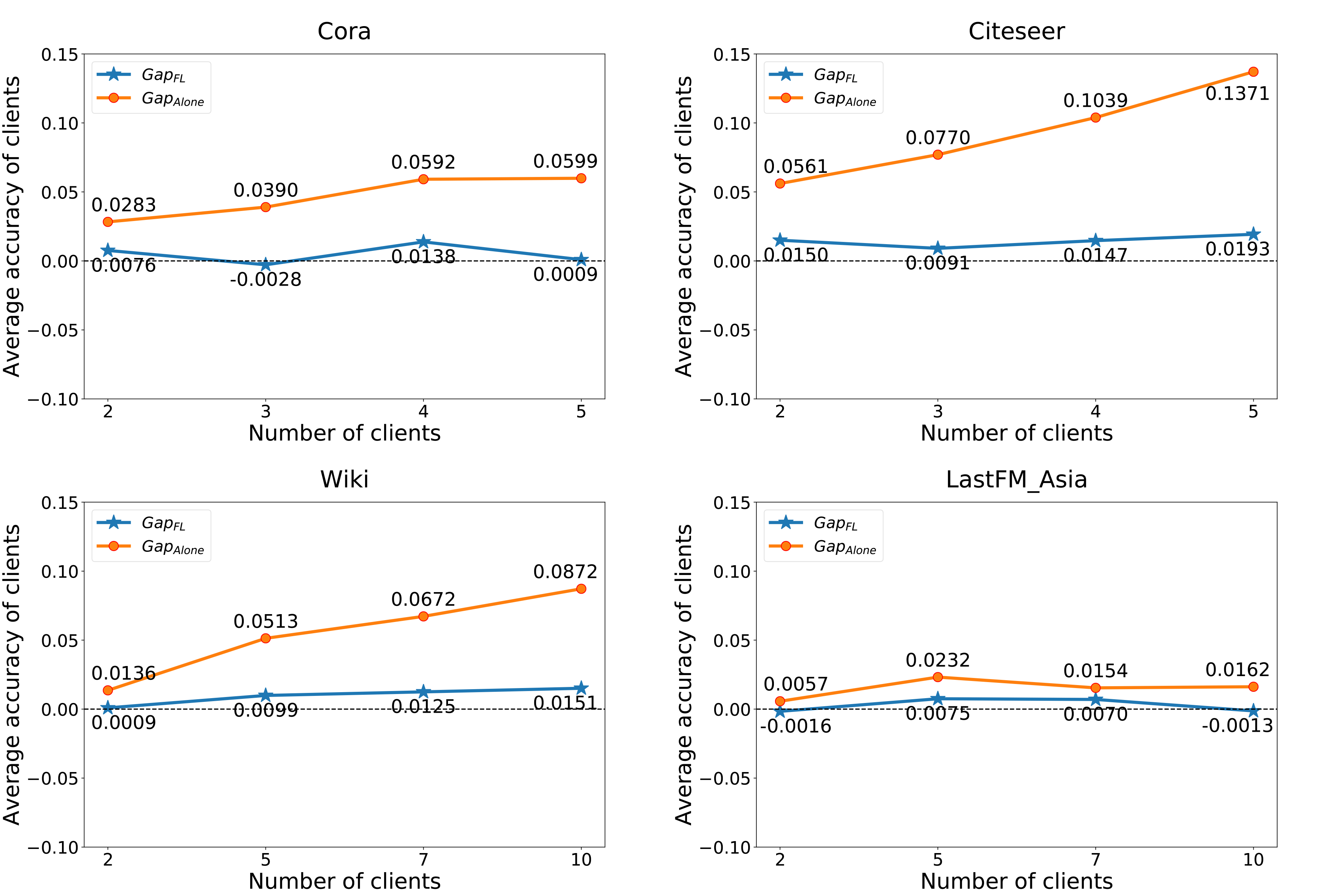} 
    \caption{Trend chart of the influence of the number of clients on federated aggregation. The abscissa is the number of clients, and the ordinate is the average accuracy of Full\_Client minus individual training and minus FLGNN}
 \label{fig5}
\end{figure}

It is noteworthy that the impact of FLGNN may not be immediately evident when a large-scale graph network dataset is partitioned into a small number of clients. Figure~\ref{fig5} illustrates the trend between the number of clients and the influence of FLGNN, where Gap FL denotes the accuracy gap between centralized and federated training models, and Gap Alone represents the accuracy gap between centralized and individual training models. As the dataset is split into more clients, each client's data volume decreases, but FLGNN can still achieve the accuracy benchmark, leading to a widening gap between individual training and FLGNN, while the gap between centralized training and FLGNN remains relatively constant. In addition, it is worth noting that the degree value of the LastFM Asia node is excessively large, such that even when it is divided into ten clients, there is still enough information available to train a model with improved accuracy, thereby rendering the effect of FLGNN not easily discernible. In summary, the overall volume and quality of data possessed by each client have a greater impact on the actual effectiveness of FLGNN than the number of clients.

\begin{table}[h]
\centering
\caption{Experimental Results}
\setlength{\tabcolsep}{1.4mm}
\centering
\resizebox{0.5\linewidth}{!}{
\begin{tabular}{cccccccc}
\hline
\multicolumn{1}{l}{} & \multicolumn{1}{l}{}       & \multicolumn{3}{c}{Cora}                   & \multicolumn{3}{c}{Citeseer}      \\ \cline{3-8} 
\multicolumn{1}{l}{} & \multicolumn{1}{l}{Client} & FLGNN           & Alone  & Full            & FLGNN  & Alone  & Full            \\ \hline
\multirow{2}{*}{2}   & A                          & 0.7990          & 0.7836 & \textbf{0.8131} & 0.6295 & 0.5898 & \textbf{0.6390} \\
                     & B                          & 0.7640          & 0.7380 & \textbf{0.7651} & 0.6321 & 0.5895 & \textbf{0.6525} \\ \hline
\multirow{3}{*}{3}   & A                          & \textbf{0.7924} & 0.7308 & 0.7907          & 0.6266 & 0.5387 & \textbf{0.6413} \\
                     & B                          & \textbf{0.7558} & 0.7337 & 0.7500          & 0.6218 & 0.5644 & \textbf{0.6303} \\
                     & C                          & \textbf{0.7707} & 0.7290 & 0.7700          & 0.6483 & 0.5897 & \textbf{0.6522} \\ \hline
\multirow{4}{*}{4}   & A                          & 0.8156          & 0.7646 & \textbf{0.8354} & 0.6257 & 0.5018 & \textbf{0.6359} \\
                     & B                          & 0.7604          & 0.7231 & \textbf{0.7771} & 0.5970 & 0.5080 & \textbf{0.6186} \\
                     & C                          & \textbf{0.7353} & 0.6885 & 0.7293          & 0.6359 & 0.5657 & \textbf{0.6533} \\
                     & D                          & 0.7661          & 0.7197 & \textbf{0.7911} & 0.6277 & 0.5539 & \textbf{0.6373} \\ \hline
\multirow{5}{*}{5}   & A                          & 0.7893          & 0.7235 & \textbf{0.8079} & 0.6226 & 0.4747 & \textbf{0.6523} \\
                     & B                          & \textbf{0.7782} & 0.7099 & 0.7685          & 0.6281 & 0.5336 & \textbf{0.6415} \\
                     & C                          & \textbf{0.7674} & 0.7147 & 0.7581          & 0.6229 & 0.5219 & \textbf{0.6283} \\
                     & D                          & 0.7689          & 0.7277 & \textbf{0.7766} & 0.6692 & 0.5577 & \textbf{0.6927} \\
                     & E                          & \textbf{0.7459} & 0.6785 & 0.7428          & 0.6502 & 0.5162 & \textbf{0.6746} \\ \hline
\end{tabular}}
\label{table4}
\end{table}

\begin{table}[h]
\setlength{\tabcolsep}{1.2mm}
\centering
\resizebox{0.5\linewidth}{!}{
\begin{tabular}{cccccccc}
\hline
                     & \multicolumn{1}{l}{} & \multicolumn{3}{c}{Wiki}                   & \multicolumn{3}{c}{LastFM Asia}           \\ \cline{3-8} 
                     & Client               & FLGNN           & Alone  & Full            & FLGNN           & Alone  & Full            \\ \hline
\multirow{2}{*}{2}   & A                    & \textbf{0.6446} & 0.6318 & 0.6441          & \textbf{0.8334} & 0.8334 & 0.8317          \\
                     & B                    & 0.6641          & 0.6516 & \textbf{0.6665} & \textbf{0.8292} & 0.8146 & 0.8277          \\ \hline
\multirow{5}{*}{5}   & A                    & 0.6475          & 0.6106 & \textbf{0.6572} & 0.8086          & 0.7992 & \textbf{0.8166} \\
                     & B                    & 0.6266          & 0.5792 & \textbf{0.6411} & 0.8177          & 0.8024 & \textbf{0.8272} \\
                     & C                    & 0.6446          & 0.6150 & \textbf{0.6504} & 0.8341          & 0.8141 & \textbf{0.8416} \\
                     & D                    & 0.6714          & 0.6264 & \textbf{0.6826} & 0.8291          & 0.8127 & \textbf{0.8331} \\
                     & E                    & 0.6489          & 0.6008 & \textbf{0.6573} & 0.8264          & 0.8091 & \textbf{0.8350} \\ \hline
\multirow{7}{*}{7}   & A                    & 0.6311          & 0.5961 & \textbf{0.6451} & 0.8202          & 0.8158 & \textbf{0.8277} \\
                     & B                    & 0.6459          & 0.5864 & \textbf{0.6606} & 0.8353          & 0.8240 & \textbf{0.8442} \\
                     & C                    & 0.6423          & 0.5843 & \textbf{0.6523} & 0.8288          & 0.8308 & \textbf{0.8341} \\
                     & D                    & 0.6771          & 0.6263 & \textbf{0.6898} & 0.8273          & 0.8114 & \textbf{0.8321} \\
                     & E                    & 0.6588          & 0.6141 & \textbf{0.6749} & 0.8242          & 0.8112 & \textbf{0.8331} \\
                     & F                    & 0.6540          & 0.6123 & \textbf{0.6663} & 0.8261          & 0.8182 & \textbf{0.8320} \\
                     & G                    & 0.6342          & 0.5414 & \textbf{0.6424} & 0.8400          & 0.8316 & \textbf{0.8477} \\ \hline
\multirow{10}{*}{10} & A                    & 0.6434          & 0.5717 & \textbf{0.6590} & 0.8352          & 0.8081 & \textbf{0.8373} \\
                     & B                    & 0,6564          & 0.5681 & \textbf{0.6645} & \textbf{0.8317} & 0.8175 & 0.8274          \\
                     & C                    & 0.6534          & 0.5830 & \textbf{0.6690} & 0.8200          & 0.8011 & \textbf{0.8204} \\
                     & D                    & 0.6209          & 0.5483 & \textbf{0.6300} & \textbf{0.8316} & 0.8184 & 0.8289          \\
                     & E                    & 0.6414          & 0.5602 & \textbf{0.6535} & \textbf{0.8315} & 0.8137 & 0.8293          \\
                     & F                    & 0.6345          & 0.5703 & \textbf{0.6528} & \textbf{0.8304} & 0.8115 & 0.8296          \\
                     & G                    & 0.6567          & 0.5846 & \textbf{0.6745} & \textbf{0.8447} & 0.8183 & 0.8419          \\
                     & H                    & 0.6305          & 0.5783 & \textbf{0.6460} & \textbf{0.8309} & 0.8147 & 0.8293          \\
                     & I                    & 0.6434          & 0.5591 & \textbf{0.6562} & 0.8333          & 0.8150 & \textbf{0.8316} \\
                     & J                    & 0.6422          & 0.5784 & \textbf{0.6687} & 0.8283          & 0.8249 & \textbf{0.8296} \\ \hline
\end{tabular}}
\end{table}

\subsection{Attack and Defense Experiments}
\begin{figure*}[]
    \centering 
    \label{Fig.sub.3}
    \includegraphics[width=1\textwidth]{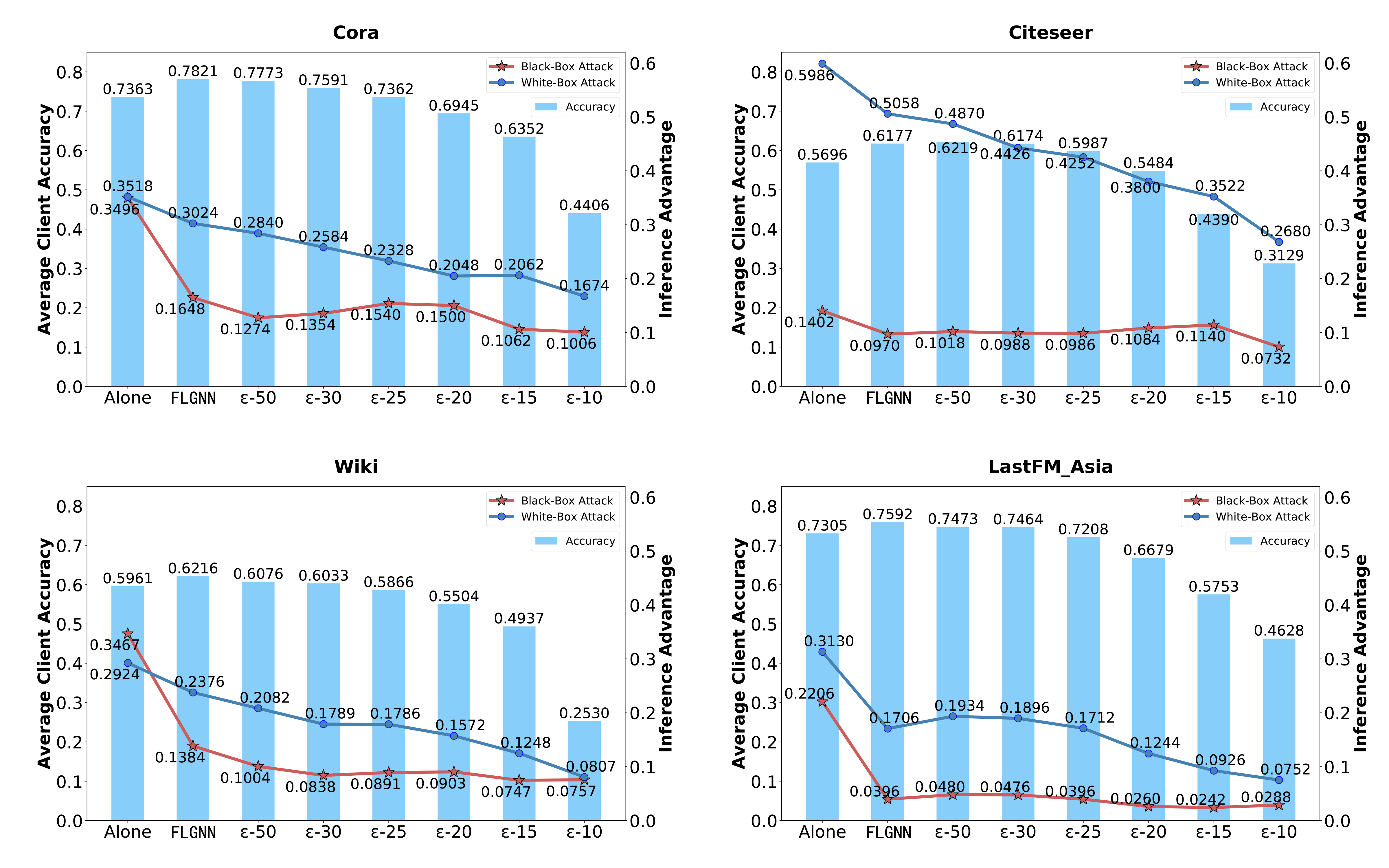}
    \caption{The bar graph represents the average accuracy of each client model under various settings, and the line graph represents the inference advantage of black box and white box attacks on a certain client.}
\label{fig6}
\end{figure*}

To evaluate the performance of FLGNN in preserving node privacy, this section conducts membership inference attacks, including white-box attacks by federation members and black-box attacks by external sources. Additionally, a differential privacy defense experiment using Laplace noise is established to examine the effect of differential privacy defense. The model obtained from experiment C is the target of the membership inference attack. In this scenario, the adversary can only infer membership affiliation based on the model output (black-box) or model intermediate parameters (white-box), and no other clients are allowed to provide further node information. All attacks and defenses are conducted under the assumption of plaintext data sharing. Since membership inference attack is a binary classification issue, random guessing accuracy can also approach 50\%, hence the ``inference advantage'' $I\_adv$ is utilized as the indicator~\cite{duddu2020quantifying}, and $I\_acc$ is the attack success rate in Eq 8. Figure~\ref{fig6} depicts the attack and defense result, where Alone's black-box and white-box attack targets are model trained alone, while the rest of the attacks are among federated learning participants. Using $\epsilon$ in the abscissa represents adding differential privacy in federated learning, and the number represents the value of the privacy budget. The higher the privacy budget, the less noise is added.

\begin{equation} \label{eq8}
\centering
I_{adv} = (I_{acc}-0.5)\times 2
\end{equation}

Figure~\ref{fig6} illustrates that FLGNN exhibits superior security compared to standalone training in most scenarios, regardless of whether under black-box or white-box attacks. This is attributed to the fact that the data is trained locally and only weight parameters are uploaded, which have minimal correlation with the node features that carry private information. Even if an internal malicious client obtains the weight parameters of other clients, identifying the precise node of the client’s training set is challenging. The addition of differential privacy further reduces the inference advantage, as well as the model's accuracy. Smaller privacy budgets offer better security, but weaken the model's performance. When excessive noise is added, FLGNN's model performance may not even be as good as standalone training, thereby rendering federated learning meaningless. Moreover, since noise is incorporated into the shared weight parameter, adversaries cannot access the weight parameter in black-box settings. Therefore, differential privacy is ineffective against black-box attacks. Interestingly, a moderate amount of noise may occasionally improve model performance, as the added noise may nudge the weight parameters towards correct classification.

\subsection{Different Types of Edges Experiments}
In this experiment, we explore a scenario in which the network that clients hold has different edges to validate the efficacy of FLGNN+. The experiments are conducted on Yelp Urbana and Terrorist Attack. For Yelp Urbana, two clients hold the friend network and the co-dining network, respectively. Then, each client intercepts the user nodes that left comments before 2018 as the training and validation set and collects the nodes that left comments after 2018 as the test set to evaluate the model's categorization of users' preferences. The Terrorist Attack dataset has previously been divided into Loc and Loc\_org. Loc denotes a terrorist attack that occurred in the same location, whereas Loc\_org denotes a terrorist attack that occurred in the same location and was carried out by the same organization. Due to the network's tiny size, the training, validation, and test sets are separated in a ratio 1:1:3. The initial aggregation weight is $\gamma_{u,0}^u=0.5$, which is the same as the average aggregation weight.
\hspace*{1em}From the results in Table~\ref{table5}, it can be seen that FLGNN+ performs better than the federated average algorithm on the two real datasets, indicating that for networks with different types of edges, the simple average aggregation will ignore the influence of the edge type on the training task. Individual training has a big accuracy gap for Yelp Urbana since the node degree of the co-dining network is substantially larger than that of the friend network. At this moment, adopting FLGNN's average aggregation approach can greatly increase the accuracy of a smaller friend network. However, the model's accuracy diminishes for the co\-dining network. When the global model obtained by the client does not perform as well on the classification task as it is trained alone, the client has no meaning to participate in federated learning. FLGNN+ is not considerably improved over FLGNN in Terrorist Attacks due to the tiny quantity of data and Loc\_org is a subgraph of Loc but it is still performing better than FLGNN.

\begin{table}[]
\caption{Experiment Results}
\setlength{\tabcolsep}{2.1mm}
\centering
\begin{tabular}{ccccc}
\hline
                                  & Network   & Alone  & FLGNN  & FLGNN+          \\ \hline
\multirow{2}{*}{Yelp Urbana}      & Friend    & 0.5337 & 0.7034 & \textbf{0.7278} \\
                                  & Co-dining & 0.7963 & 0.7778 & \textbf{0.8056} \\ \hline
\multirow{2}{*}{Terrorist Attack} & Loc       & 0.6926 & 0.6915 & \textbf{0.6950} \\
                                  & Loc\_org  & 0.7112 & 0.7180 & \textbf{0.7363} \\ \hline
\end{tabular}
\label{table5}
\end{table}

\section{Conclusion and Future Work} \label{conclusion}
This work proposes a federated aggregation method FLGNN based on the graph neural network for node classification. In scenarios with varied node and graph structures, federated aggregation is performed by sharing the weight parameters of multiple graph neural network layers to obtain a model combined with performance and safety. The experiments using citation networks and social networks prove that FLGNN is suitable for various horizontal federation scenarios, and has a certain defensive effect on membership inference attacks from external and internal clients. For scenarios with different edge types in graph networks, this paper designs a dynamic weight aggregation mechanism with feedback based on FLGNN and verifies its effectiveness on real data sets. Furthermore, we find that when the number of clients with smaller data achieves a certain quantity, FLGNN can gather information from multiple parties to train an excellent global model, which makes some edge devices with a small amount of data values. However, edge devices are often limited by communication bandwidth, affecting federated aggregation's efficiency. Although this paper does not go into great length about the communication efficiency of federated learning, this area should be researched more.

\subsubsection*{Acknowledgments.} 
This work was supported in part by the Key R\&D Program of Zhejiang under Grants 2022C01018 and 2024C01025, by the National Natural Science Foundation of China under Grants 62103374,U21B2001 and 61973273 and by the Key R\&D Projects in Zhejiang under Grant 2021C01117.
Hao Song and Jiacheng Yao contribute equally to this work.

\bibliographystyle{splncs04_}
\bibliography{sample-base}

\end{document}